\documentclass[conference]{IEEEtran}
\IEEEoverridecommandlockouts
\usepackage{cite}
\usepackage{amsmath,amssymb,amsfonts}
\usepackage{algorithmic}
\usepackage{graphicx}
\usepackage{textcomp}
\usepackage{xcolor}
\usepackage{hyperref}
\def\BibTeX{{\rm B\kern-.05em{\sc i\kern-.025em b}\kern-.08em
    T\kern-.1667em\lower.7ex\hbox{E}\kern-.125emX}}
\begin{document}

\title{
SaENeRF: Suppressing Artifacts in Event-based Neural Radiance Fields
}

\author{
\IEEEauthorblockN{
Yuanjian Wang\textsuperscript{1},
Yufei Deng\textsuperscript{1},
Rong Xiao\textsuperscript{1*},
Jiahao Fan\textsuperscript{1},
Chenwei Tang\textsuperscript{1},
Deng Xiong\textsuperscript{2},
and Jiancheng Lv\textsuperscript{1*}
}
\IEEEauthorblockA{
\textsuperscript{1}College of Computer Science, Sichuan University, Chengdu, China
}
\IEEEauthorblockA{
\textsuperscript{2}Stevens Institute of Technology, Hoboken, USA
}
\thanks{*Corresponding authors: \{rxiao, lvjiancheng\}@scu.edu.cn}
% \thanks{The code is available at \href{https://github.com/Mr-firework/SaENeRF}{https://github.com/Mr-firework/SaENeRF}.}
}

\maketitle

\begin{abstract}
Event cameras are neuromorphic vision sensors that asynchronously capture changes in logarithmic brightness changes, offering significant advantages such as low latency, low power consumption, low bandwidth, and high dynamic range. While these characteristics make them ideal for high-speed scenarios, reconstructing geometrically consistent and photometrically accurate 3D representations from event data remains fundamentally challenging. Current event-based Neural Radiance Fields (NeRF) methods partially address these challenges but suffer from persistent artifacts caused by aggressive network learning in early stages and the inherent noise of event cameras.
To overcome these limitations, we present SaENeRF, a novel self-supervised framework that effectively suppresses artifacts and enables 3D-consistent, dense, and photorealistic NeRF reconstruction of static scenes solely from event streams.
Our approach normalizes predicted radiance variations based on accumulated event polarities, facilitating progressive and rapid learning for scene representation construction. Additionally, we introduce regularization losses specifically designed to suppress artifacts in regions where photometric changes fall below the event threshold and simultaneously enhance the light intensity difference of non-zero events, thereby improving the visual fidelity of the reconstructed scene.
Extensive qualitative and quantitative experiments demonstrate that our method significantly reduces artifacts and achieves superior reconstruction quality compared to existing methods. The code is available at \url{https://github.com/Mr-firework/SaENeRF}.
\end{abstract}


\section{Introduction}
\label{sec:intro}

Event cameras \cite{gallego2020event_survey} are highly promising bio-inspired neuromorphic vision sensors \cite{zhu2024spikenerf, guo2024spike}.
Unlike frame-based cameras, which capture a sequence of light intensities through synchronized exposure, event cameras asynchronously generate continuous events by detecting changes in brightness that exceed a preset threshould. 
This unique mechanism provides several distinctive advantages, including low latency, low power consumption, low bandwidth, and high dynamic range. 

However, reconstructing both geometry and photometry from event stream remains a significant challenge. Existing event-based visual odometry methods \cite{kim2016real-time_3d_reconstruct_tracking, Zhou18eccvSemi-dense_3D, Zhou21tro-esvo}, inspired by traditional frame-based approaches such as Kalman filters \cite{mourikis2007multi}, typically generate sparse or semi-dense point clouds. These reconstructions lack absolute light intensity information due to the inherent characteristics of event cameras. On the other hand, methods like E2VID \cite{Rebecq19pamie2vid}, which aim to reconstruct photometric images directly from events, heavily rely on pre-training using simulated data \cite{rebecq2018esim}. Although promising, these approaches often struggle with accuracy and consistency, primarily because they do not fully leverage geometric consistency in the reconstruction process.

% Against this backdrop, 
Neural Radiance Fields (NeRF) \cite{mildenhall2020nerf} have emerged as a groundbreaking approach. NeRF introduces an innovative methodology to represent scenes by implicitly modeling geometry and photometry through continuous volumetric representations. This paradigm shift offers the potential to overcome the limitations of traditional event-based methods, enabling dense, photorealistic reconstructions with both geometric and photometric consistency.
Recently, studies like EventNeRF \cite{rudnev2023eventnerf}, E-NeRF \cite{klenk2023e-nerf}, and others \cite{low2023_robust-e-nerf, Hwang_2023_WACV}, have reconstructed NeRF using purely event-based supervision. They rely on alternative reconstruction losses inspired by or derived from Event Generation Model (EGM) \cite{gallego2020event_survey, hidalgo22event-aided_direct_sparse_odometry}, aiming to overcome the limitations of standard camera and further bridge the gap between event stream and high-quality scene reconstruction.

However, a significant challenge in event-based NeRFs \cite{rudnev2023eventnerf, Hwang_2023_WACV, low2023_robust-e-nerf, wang2024physical} persists: artifacts appearing in the background and on objects continue to degrade the overall quality of scene reconstruction.
\textit{A natural problem arises: why do traditional NeRFs not exhibit significant artifact issues under ideal conditions, whereas event-based NeRFs do?} 
A key reason lies in their supervision mechanisms. Traditional NeRFs utilize intensity information for direct supervision, which efficiently suppresses artifacts during training. In contrast, event-based NeRFs rely solely on light intensity differences, which are less effective and can lead to artifact propagation during training, making artifact removal more challenging. These artifacts often originate from misrepresentations in the early stages of training that are not promptly corrected. Additionally, they are exacerbated by the inherent background noise of event cameras.
Although methods such as \cite{rudnev2023eventnerf, klenk2023e-nerf} has endeavored to suppress artifacts by increasing the sampling of pixels with zero accumulated event polarities (referred to as zero-events) and leveraging the inconsistency of artifacts across multiple views, these efforts have not fully resolved the issue. 
The key to effective artifact suppression lies in avoiding the learning of excessively large variations in light intensity during the initial stages of model training, prior to geometry reconstruction, and developing a more effective approach to mitigate artifacts in zero-events.

In this paper, we propose SaENeRF to suppress artifacts and achieve high-fidelity reconstructions, thereby enabling 3D-consistent, dense, and photorealistic NeRF reconstructions of static scenes directly from data captured by a moving event camera.
Firstly, we introduce the construction loss function which employs the L1-norm to normalize the accumulated event polarities as well as the predicted magnitude of photometric changes, with the aim of facilitating the progressive learning of light difference.
To boost learning efficiency and hasten model convergence, we normalize only the components where the predicted photometric changes are aligned with the corresponding accumulated event polarities. 
Nevertheless, this is inadequate to eliminate artifacts, let alone when combined with the inherent background noise of the event camera.
We further introduce zero-events regularization losses, which are capable of suppressing zero-events photometric changes and simultaneously boosting the light difference of non-zero events.
We qualitatively and numerically evaluate our techniques for the novel view synthesis task using the dataset proposed by Rudnev et al. \cite{rudnev2023eventnerf}, including synthetic and real event sequences. 
Our experiments, compared with recent methods \cite{rudnev2023eventnerf, klenk2023e-nerf}, show that our approach effectively alleviates artifacts and achieves better results.

\section{Related work}
\paragraph{Event-based 3D Reconstruction} 
The field of 3D reconstruction technology for event cameras has witnessed substantial advancements and can be broadly delineated based on distinct technical methodologies.
The first category encompasses the fusion of multimodal data, including depth sensors \cite{zuo2022devo}, standard cameras \cite{kueng2016low, cho2022event}, LiDAR \cite{gehrig2021dsec, cui2022dense}, and IMUs \cite{vidal2018ultimate, muglikar2021calibrate}, to facilitate the reconstruction process. These methodologies leverage the combined strengths of various sensors to enhance the precision and completeness of the reconstructions. However, they often fail to fully capitalize on the unique capabilities and potential inherent solely in event cameras.
Other approaches focus exclusively on utilizing event cameras for SLAM or visual odometry, whether through the deployment of a single event camera \cite{kim2016real-time_3d_reconstruct_tracking} or a stereo pair of event cameras \cite{Zhou18eccvSemi-dense_3D, Zhou21tro-esvo}. These techniques typically yield sparse point clouds or achieve semi-dense 3D reconstructions, primarily emphasizing features like edges and corners.
While \cite{xiao2022event} managed to achieve dense 3D reconstructions by applying traditional Structure from Motion (SFM) and Multi-View Stereo (MVS) techniques on recovered video frames derived from events \cite{Rebecq19pamie2vid}, the accuracy of the reconstructed scenes still leaves much to be desired.

\paragraph{Event-based Neural Radiance Fields} 
Recent advancements in neural implicit reconstruction have merged Neural Radiance Fields (NeRF) \cite{mildenhall2020nerf} with Event Generation Model (EGM) \cite{gallego2020event_survey, hidalgo22event-aided_direct_sparse_odometry}.
Methods such as EventNeRF \cite{rudnev2023eventnerf}, Ev-NeRF \cite{Hwang_2023_WACV} and E-NeRF \cite{klenk2023e-nerf} have pioneered in reconstructing dense and consistent 3D implicit representations of scenes directly from the event stream, optionally with the assistance of RGB data \cite{klenk2023e-nerf}. 
However, these reconstructions frequently exhibit artifacts. 
Robust e-NeRF \cite{low2023_robust-e-nerf} deeply explores the intrinsic parameters of event cameras to enhance robustness in scenarios with sparse and noisy events generated under non-uniform motion.
Meanwhile, PAEv3d \cite{wang2024physical} incorporates physical constraints, derived from event data priors, to enhance the training process of NeRF. Nonetheless, artifacts still persist and have not been adequately explored.
By utilizing both RGB and event, E2NeRF \cite{qi2023e2nerf}, EvDeblurNeRF \cite{cannici2024mitigating} and DE-NeRF \cite{ma2023deformable} have achieved remarkable results in mitigating motion blur \cite{cannici2024mitigating} and deformable neural radiance fields \cite{bhattacharya2024evdnerf}. The integration with an RGB camera has helped to mitigate the challenges associated with artifacts removal, although color artifacts still sporadically appear.
EN-SLAM \cite{qu2023implicit}, the first event-RGBD implicit neural SLAM framework, effectively leverages the event stream and RGBD data to overcome challenges posed by extreme motion blur and varying lighting conditions, without requiring known poses. This multi-modal approach effectively alleviates the artifacts resulting from the event data. 
The integration of 3D Gaussian Splatting \cite{kerbl20233d} within Event3DGS \cite{xiong2024event3dgs} results in rendering speeds that surpass those of NeRF-based methods. However, artifacts continue to be present and require further investigation.

\section{Method}
\label{method}

\begin{figure}[tp]
    \includegraphics[width=\linewidth]{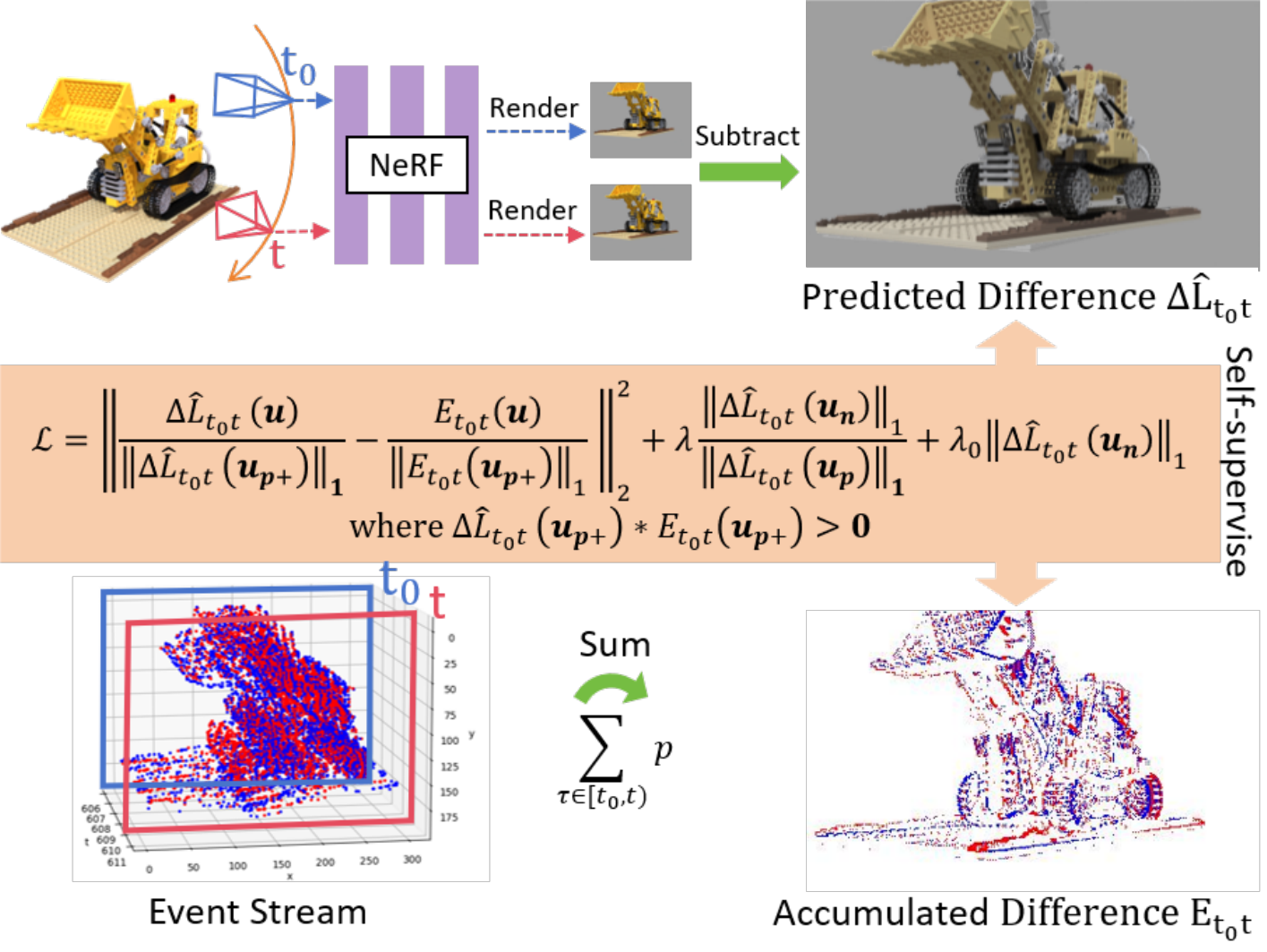}
    \caption{The overview presents our progressive learning method. By normalizing predicted radiance variations based on accumulated event polarities, our approach facilitates progressive and rapid training. Furthermore, recognizing the inconsistency of artifacts across multiple views, we incorporate zero-event regularization losses to suppress artifacts in zero accumulated event polarities, thereby enhancing the overall outcome.}
    \label{fig:pipeline}
\end{figure}

Given an input continuous event stream $\mathcal{E}$ captured by a moving event camera, our goal is to leverage Neural Radiance Fields (NeRF) \cite{mildenhall2020nerf} to learn an implicit neural 3D scene representation $F_\theta$, while effectively mitigating artifacts to construct a high-fidelity 3D representation of a static scene. 
We assume that the event camera intrinsics $\mathbf{K}\in\mathbb{R}^{3\times3}$ and camera poses $\mathbf{P}\in {SE}(3)$ are known. 
The overview of our method is presented in Fig.\ref{fig:pipeline}.
We adopt NeRF as our 3D scene representation due to its exceptional performance in novel view synthesis. To formulate a self-supervised loss function for our event-based NeRF, we introduce an Event Generation Model (EGM). Building on this, we analyze the proposed event normalization loss and present a series of optimization techniques designed to reduce artifacts and facilitate progressive learning. However, relying solely on event normalization loss is insufficient to eliminate artifacts caused by event noise. To address this, we introduce zero-events regularization losses, which impose stronger constraints to further suppress artifacts. Finally, we provide a detailed description of the implementation.

\subsection{Preliminaries: neural radiance fields}
\label{sec:nerf}

Neural Radiance Fields (NeRF) \cite{mildenhall2020nerf} reconstruct and render 3D scenes from a set of 2D images, capturing both the geometry and appearance of the scene to enable photorealistic novel view synthesis. The core of NeRF lies in its exploitation of a Multi-Layer Perceptron (MLP), denoted by $F_\Theta:(\mathbf{x}, \mathbf{d})\rightarrow{}(\sigma, \mathbf{c})$, to implicitly represent the scene as a continuous 5D function that outputs the color $\mathbf{c}=(r, g, b)$ and density $\sigma$ at any given 3D location $\mathbf{x}=(x,y,z)$ and viewing direction $\mathbf{d}=(\theta, \phi)$. 
The scene representation $F_\Theta$ is optimized by minimizing the difference between the rendered pixel colors and the ground truth pixel colors from a set of captured images with known camera parameters.
Specifically, to render a pixel, a ray $\bf{r}$ is cast from the camera center $\bf{o}$ through the pixel in the image plane along its normalized view direction $\mathbf{d}$. A set of points $\{\mathbf{x}_i|\mathbf{x}_i = \mathbf{o} + t_i \mathbf{d} \}_{i=1}^S$ are then sampled along the ray $\bf{r}$, where $t_i$ represents the distance from the camera center $\bf{o}$ to the sample point $\mathbf{x}_i$, and $S$ denotes the number of 3D points sampled along the ray. The final rendered color $\mathbf{I}$ for the ray $\bf{r}$ is calculated using volume rendering \cite{max1995optical, kajiya1984ray} as:
\begin{equation}
\begin{split}
\mathbf{I}(\mathbf{r})=\sum_{i=1}^S{T_i\alpha_i\mathbf{c_i}},
\text{ where } T_i=\prod_{j=1}^{i-1}(1-\alpha_j),\\ 
\alpha_i = 1 - \exp (-\sigma_i\delta_i),
\label{eq:volume_rendering}
\end{split}
\end{equation}
where $T_i$ represents the transmittance, $\alpha_i$ indicates the opacity for the sample point $\mathbf{x}_i$, and $\delta_i$ is the distance between neighboring sample points.

\subsection{Preliminaries: event generation model}
\label{sec:EGM}

NeRF directly utilizes the color intensity obtained from the rendering process, along with the original input images, to define its loss function in a self-supervised manner. 
In contrast, event-based NeRF primarily relies on Event Generation Model (EGM) \cite{gallego2020event_survey, hidalgo22event-aided_direct_sparse_odometry} to establish a connection between the rendered light intensity differences and the event stream $\mathcal{E}=\{\mathbf{e}|\mathbf{e}=(\mathbf{u}, t, p)\}$. 
Specifically, an event $\mathbf{e}$ is triggered at pixel $\mathbf{u}$ when the detected change in log-radiance exceeds the event threshold $C$. 
This EGM can be represented by the following equation:
\begin{equation}
    \begin{split}
    p C = \log\mathbf{I}(\mathbf{u},t)-\log\mathbf{I}(\mathbf{u},t^{prev}), \text{ where } C > 0,
    \end{split}
    \label{eq:event_def}
\end{equation}
where the event polarity $p \in$ \{+1,-1\} indicates whether the brightness at pixel $\bf{u}$ has increased or decreased from the previous event triggered time $t^{prev}$ to time $t$.

For efficiency and simplicity, our approach utilizes event windows \cite{rudnev2023eventnerf} rather than processing individual events\cite{low2023_robust-e-nerf, klenk2023e-nerf, cannici2024mitigating}. 
And the event windows, defined over a suitable time interval $\tau$, facilitate the convergence of training.
Furthermore, the log-light difference $\Delta L_{t_0t}(\mathbf{u}) = \mathrm{log}\mathbf{I}(\mathbf{u},t) - \mathrm{log}\mathbf{I}(\mathbf{u},t_0) $ at pixel $\mathbf{u}$ in temporal window $\tau \in [t_0, t)$
% from $t_0$ to $t$ 
can be estimated using the accumulated event polarities $E_{t_0t}(\mathbf{u})$ multiplied by the event threshold $C$, as follows:
\begin{equation}
\begin{split}
\Delta L_{t_0t}(\mathbf{u})=\sum_{\tau\in[t_0,t)} pC\overset{def}{=}E_{t_0t}(\mathbf{u})C .
\end{split}
\label{eq:EGM}
\end{equation}

\subsection{Event normalization loss}
\label{sec:Event Norm Loss}

\begin{figure*}[tp]
    \centering
    \includegraphics[width=.9\linewidth]{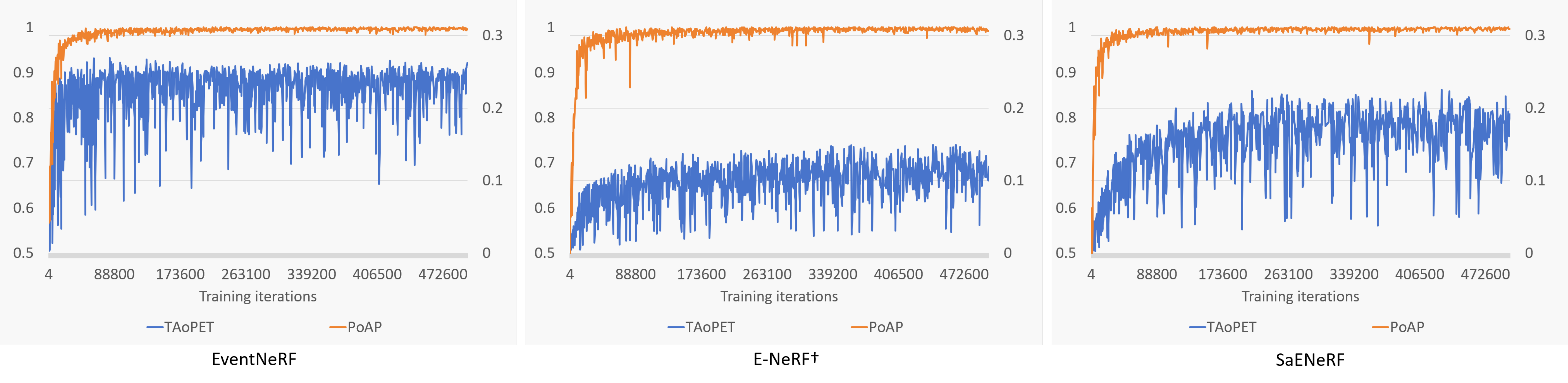}
    \caption{EventNeRF adopts an aggressive joint optimization strategy, simultaneously learning in geometric structure via PoAP and light difference via TAoPET, but often introduces artifacts during early training phases. In contrast, E-NeRF\dag{} and SaENeRF match EventNeRF's geometric recovery speed while adopting a more cautious light difference learning approach. However, E-NeRF\dag{}'s no-event loss \cite{klenk2023e-nerf} leads to reduced contrast and incomplete artifact suppression. }
    \label{fig:abla_lines}
\end{figure*}

The explicit event threshold setting employs an aggressive training strategy, which often lead to premature learning of scene illumination differences before the convergence of geometric structures during the initial training phase as EventNeRF in Fig.\ref{fig:abla_lines}. This premature learning could result in the generation of substantial artifacts, which not only pose challenges for subsequent removal but also risk propagating through the training process, thereby exacerbating their persistence and complicating their elimination.

Drawing inspiration from \cite{klenk2023e-nerf}, we have identified that normalization techniques can effectively prioritize geometric structure restoration in the early training stages while deferring the acquisition of illumination variations as E-NeRF\dag{} and SaENeRF in Fig.\ref{fig:abla_lines}. The fundamental principle of this approach is to direct the model's primary focus toward geometric structure learning when image contrast remains relatively low during initial training. As training progresses, the model gradually develops the capability to simultaneously learn both geometric structures and illumination differences in a coordinated manner.

To address the challenge of learning outlier illumination variations during the initial training phase, we implement the L1-norm as a replacement for the L2-norm. This strategic choice is motivated by the L1-norm's inherent robustness to outliers and its tendency to mitigate the influence of light difference, thereby enabling the model to focus primarily on acquiring dominant geometric structures without excessive interference from intricate color variations in the early training stages. At each pixel $\mathbf{u}$, we establish a normalized connection between the accumulated event polarities $E_{t_0t}(\mathbf{u})$ and the rendered illumination differences $\Delta \hat{L}_{t_0t}(\mathbf{u})$. The mathematical formulation of this connection is established through the following loss function:
\begin{equation}
\mathcal{L}_{\text{norm}}(t_0, t) = \Vert \frac{\Delta \hat{L}_{t_0t}(\mathbf{u})}{ \Vert \Delta \hat{L}_{t_0t}(\mathbf{u}) \Vert_1}-\frac{E_{t_0t}(\mathbf{u})}{\Vert E_{t_0t}(\mathbf{u})\Vert_1} \Vert_2^2.
\label{eq:EGM_norm}
\end{equation}

However, our experimental analysis reveals that the normalization approach may inadvertently introduce reconstruction artifacts. This phenomenon originates from the norm distance computation when jointly processing both: positive sampling pixels $\mathbf{u}_{p}$ with non-zero accumulated event polarities, and negative sampling pixels $\mathbf{u}_{n}$ exhibiting zero event polarities. 
While negative sampling demonstrates crucial artifact suppression capabilities as documented in \cite{rudnev2023eventnerf}, we identify an inherent conflict in photometric consistency: when $E_{t_0t}(\mathbf{u}_{n})\equiv0$ while $\Delta \hat{L}_{t_0t}(\mathbf{u}_{n})\not\equiv0$, the induced variation $\Vert\Delta \hat{L}_{t_0t}(\mathbf{u})\Vert_1$ 
becomes non-physically amplified relative to the actual event variation  $\Vert E_{t_0t}(\mathbf{u}_p)\Vert_1$. 
This discrepancy introduces secondary noise components into the reconstruction pipeline, ultimately degrading the signal-to-noise ratio in synthesized imagery. 
To resolve this fundamental conflict, we propose a constrained normalization scheme that exclusively considers positive sampling regions. The refined loss function is formulated as:
\begin{equation}
\mathcal{L}_{\text{norm-}}(t_0, t) = \Vert \frac{\Delta \hat{L}_{t_0t}(\mathbf{u})}{ \Vert \Delta \hat{L}_{t_0t}(\mathbf{u}_{p}) \Vert_1}-\frac{E_{t_0t}(\mathbf{u})}{\Vert E_{t_0t}(\mathbf{u}_{p})\Vert_1} \Vert_2^2.
\label{eq:EGM_norm_pos}
\end{equation}
 
Furthermore, building on the physical constraint that realistic event thresholds must satisfy $C>0$, we enhance our normalization framework to better align with event camera physics and improve robustness for real-world applications. This constraint originates from the inherent noise floor of event sensors, where detectable brightness variations must exceed a minimum threshold to trigger events. To operationalize this principle, we propose the \textit{Temporal Average of Predicted Event Thresholds} (TAoPET) for positive sampling pixels $\mathbf{u}_p$ over temporal window $\tau\in[t_0,t)$:
\begin{equation}
\begin{split}
    \hat{C}_{t_0t}(\mathbf{u}_p) = \frac{\Delta \hat{L}_{t_0t}(\mathbf{u}_{p})}{E_{t_0t}(\mathbf{u}_p)},
\end{split}
\label{eq:pred_C}
\end{equation}
where $\Delta \hat{L}_{t_0t}$ denotes predicted luminance variation and $E_{t_0t}(\mathbf{u}_p)$ represents accumulated event polarities. 
The TAoPET metric offers a well-founded approach for monitoring the status of illumination restoration.

We further identify \textit{photometrically consistent pixels} $\mathbf{u}_{p+}$ where predicted luminance variations $\Delta \hat{L}_{t_0t}(\mathbf{u}_{p})$ have the same sign as the accumulated event polarities $E_{t_0t}(\mathbf{u}_p)$. To effectively observe the geometric convergence, we introduce the \textit{Proportion of Appropriate Pixels} (PoAP), defined as:
\begin{equation}
\begin{split}
    \hat{\mathcal{P}}=\frac{l}{N},
\end{split}
\label{eq:proportion_C}
\end{equation}
where $l$ represents the number of photometrically consistent pixels $\mathbf{u}_{p+}$, and $N$ represents the total sampled pixels.

Consequently, we ultimately derive the normalization loss by integrating the photometrically consistent pixels in the following way:
\begin{equation}
\begin{split}
\mathcal{L}_{\text{norm+}}(t_0, t) &= \Vert
    \frac{\Delta \hat{L}_{t_0t}(\mathbf{u})}
         { \Vert \Delta \hat{L}_{t_0t}(\mathbf{u}_{p+}) \Vert_1}- 
    \frac{E_{t_0t}(\mathbf{u})}
         { \Vert E_{t_0t}(\mathbf{u}_{p+}) \Vert_1}
    \Vert_2^2.
\end{split}
\label{eq:loss_norm+}
\end{equation}

\subsection{Zero-events regularization losses}
\label{sec:Zero Events Regularization} 
% Furthermore, we found that solely relying on the normalization loss (Eq. \ref{eq:loss_norm+}) is insufficient to effectively handle noisy events caused by background noise in event cameras as norm+ illustrated in Fig. \ref{fig:abla_real}.
Spurious measurements from event cameras constitute another significant source of artifacts that degrade reconstruction quality.
To mitigate these artifacts by leveraging their inconsistency across multiple views, we introduce additional constraints on negative sampling pixels exhibiting near-zero log-intensity changes. The negative zero-events loss is formulated:
\begin{equation}
\begin{split}
\mathcal{L}_{\text{zero-}}(t_0, t)&= {\Vert \Delta \hat{L}_{t_0t}(\mathbf{u}_{n})\Vert_1}. \\
\end{split}
\label{eq:loss_zero-}
\end{equation}

While $\mathcal{L}_{\text{zero-}}$ is designed to suppress artifacts, we observe that its unconstrained application can inadvertently reduce image contrast without fully eliminating artifacts. This occurs because an overly strong $\mathcal{L}_{\text{zero-}}$ penalty may excessively suppress intensity variations, leading to low-contrast renderings while residual artifacts persist as E-NeRF\dag{} in Fig.\ref{fig:abla_lines}.
To address this limitation, we propose an enhanced loss formulation that incorporates predicted light intensity from positive sampling pixels in the denominator. This modification establishes a dynamic balance between artifact suppression and contrast preservation, enabling the reconstruction process to simultaneously achieve high-contrast rendering and effective artifact removal at negative sampling locations, thereby significantly improving overall reconstruction quality. 
The positive zero-events loss function is defined as follows:
\begin{equation}
\begin{split}
\mathcal{L}_{\text{zero+}}(t_0, t)&=\frac { \Vert \Delta 
 \hat{L}_{t_0t}(\mathbf{u}_{n}) \Vert_1 } { \Vert \Delta 
 \hat{L}_{t_0t}(\mathbf{u}_{p})\Vert_1}.
\label{eq:loss_zero+}
\end{split}
\end{equation}

To establish a comprehensive optimization framework that enhances event data processing and final rendering quality, we formulate a composite loss function $\mathcal{L}_{\text{norm+\&zero}}$ integrating three critical components:
\begin{equation}
\begin{split}
\mathcal{L}_{\text{norm+\&zero}}(t_0, t)=\mathcal{L}_{\text{norm+}}(t_0, t)&+\lambda\mathcal{L}_{\text{zero+}}(t_0, t) \\ &+\lambda_0\mathcal{L}_{\text{zero-}}(t_0, t), 
\label{eq:loss_norm+_zero-_zero+}
\end{split}
\end{equation}
where $\lambda, \lambda_0$ are hyperparameters controlling the trade-off between geometric fidelity and artifact suppression.

For temporal aggregation across multiple event windows, we design a multi-window optimization scheme:
\begin{equation}
\begin{split}
    \mathcal{L}=\frac{1}{\mathcal{M}}\sum_{i=1}^{\mathcal{M}}\mathcal{L}_{\text{norm+\&zero}}(t_0, t), \\
    t_0\sim U[\max(0, t-L_{\text{max}}), t),
    \label{eq:final loss}
\end{split}
\end{equation}
where $U$ represents uniform distribution, $\mathcal{M}$ represents the number of event windows, and $L_{\text{max}}$, as discussed in \cite{rudnev2023eventnerf}, represents the maximum time length of all event windows.

\subsection{Implementation details}
\label{sec:Implementation Details}
All experiments were conducted on a NVIDIA 3090 GPU. The model training took approximately fourteen hours for $5\cdot10^5$ iterations based on the code of EventNeRF \cite{rudnev2023eventnerf}. Our real-time implementation (SaENeRF NGP), which is based on the Instant-NGP \cite{muller2022instantngp} method of NeRFStudio \cite{tancik2023nerfstudio}, took around one minute to converge when performing $5\cdot10^3$ iterations.
The learning rates were assigned to $2\cdot10^{-4}$ and $1\cdot10^{-2}$ for SaENeRF and SaENeRF NGP respectively. In our real-time implementation, a fixed negative sampling ratio of 0.05 was applied to both synthetic and real scenes. 
The weights of the zero-events regularization loss were set as follows: $\lambda$ was set to 0.5 in SaENeRF and to 1.0 in SaENeRF NGP, while $\lambda_0$ was set to 0 in SaENeRF and to 0.5 in SaENeRF NGP.

\section{Experiments}
\label{experiments}
Both quantitative and qualitative evaluations are presented to demonstrate the effectiveness of SaENeRF in suppressing artifacts and achieving high-fidelity reconstructions. These evaluations highlight SaENeRF's capability to enable 3D-consistent, dense, and photorealistic NeRF reconstructions of static scenes directly from event-based data captured by moving cameras.
To assess our approach's validity, we specifically utilize the task of \textit{Novel View Synthesis (NVS)}. 
We selected the dataset from Rudnev et al. \cite{rudnev2023eventnerf}, which comprises synthetic and real sequences captured by a 360°-rotating color event camera around objects. The real sequences were recorded specifically under low lighting conditions. 
We chose this dataset due to its solid-colored background, which doesn't trigger valid events in the event camera, highlighting any generated artifacts.
Additionally, we applied the Bayer filter as employed in \cite{rudnev2023eventnerf} to restore color images, making artifacts more easily captured by human eyes.
Quantitative and qualitative evaluations are performed on synthetic sequences. Qualitative assessments are also carried out on real sequences, further highlighting our method's ability to mitigate artifacts caused by noisy events. Finally, we conduct ablation studies to investigate the significance of various components in our approach.

\paragraph {Baseline}
We benchmarked our method against recent works: EventNeRF \cite{rudnev2023eventnerf}, E-NeRF \cite{klenk2023e-nerf}, and a naive baseline combining E2VID \cite{Rebecq19pamie2vid} with NeRF \cite{mildenhall2020nerf}. For fairness, we conducted these comparisons under controlled settings and metrics. We reimplemented E-NeRF (E-NeRF\dag{}) using the NeRF++ framework \cite{zhang2020nerf++} from EventNeRF. We also adapted the real-time EventNeRF (EventNeRF NGP) using Instant-NGP \cite{muller2022instantngp} in NeRFStudio \cite{tancik2023nerfstudio}, yielding modified versions: EventNeRF NGP* and E-NeRF*. 

\paragraph {Metrics} 
We uniformly use a set of metrics to evaluate our method in all synthetic sequences. Specifically, we assess the reconstruction quality of gamma-corrected \cite{rudnev2023eventnerf} synthesized novel views using \textit{Peak Signal-to-Noise Ratio} (PSNR), \textit{Structural Similarity Index Measure} (SSIM) \cite{image_quality_assessment}, and AlexNet-based \textit{Learned Perceptual Image Patch Similarity} (LPIPS) \cite{Zhang_2018_CVPR_LPIPS}, comparing them to target novel views. 

\subsection{Synthetic sequences}
\label{subsec:synthetic}

\begin{table*}
\caption{Quantitative comparison of our method against EventNeRF, E-NeRF\dag{} and E2VID+NeRF.}
\label{tab:synthe}
\centering\setlength{\tabcolsep}{1.3mm}
\begin{tabular}{c c c c c c c c c c c c c}
\hline
 ~ & \multicolumn{3}{c}{E2VID+NeRF} & \multicolumn{3}{ c}{EventNeRF } & \multicolumn{3}{ c}{ E-NeRF\dag{} } &\multicolumn{3}{ c}{SaENeRF}\\
% \hline
 Scene & PSNR↑ & SSIM↑ & LPIPS↓ & PSNR↑ & SSIM↑ & LPIPS↓ & PSNR↑ & SSIM↑ & LPIPS↓ & PSNR↑ & SSIM↑ & LPIPS↓ \\
 \hline
 Drums &19.71&0.85&0.22&27.43&0.91& \bf 0.07 & 27.92 & 0.91& 0.09 & \bf 28.02 & \bf 0.92 & \textbf{0.07} \\
 Lego  &20.17&0.82&0.24&25.84&0.89& 0.13 & \bf 28.09 & 0.90 & 0.09 & 27.88 & \textbf{0.92} & \textbf{0.07} \\
 Chair &24.12&0.92&0.12&30.62&0.94& 0.05 & 30.38 & 0.94 & 0.05 & \bf 31.04 & \textbf{0.95} & \textbf{0.04} \\
 Ficus &24.97&0.92&0.10&31.94&0.94& 0.05 & \bf 32.03 & 0.94 & 0.05 & {32.01} & \textbf{0.95} & \textbf{0.04} \\
 Mic   &23.08&0.94&0.09&31.78&0.96& \textbf{0.03} & 31.95 & 0.96 & 0.05 &\textbf{32.41} & \textbf{0.97} & \textbf{0.03} \\
 Hotdog &24.38&0.93&0.12&30.26&0.94 & \textbf{0.04} & 30.29 & 0.94 & \bf0.04 &\textbf{31.28} & \textbf{0.95} & {\bf 0.04} \\
 Materials &22.01&0.92&0.13&24.10&0.94 & 0.07 & \bf 30.37 & \bf0.95 & 0.06 & {30.11} & \bf{0.95} & \textbf{0.05} \\
 \hline
 Average &22.64&0.90&0.15&28.85& 0.93 & 0.06 & 30.15 & \bf 0.94 & 0.06 & \bf{30.39} & \bf{0.94} & \bf{0.05} \\
 \hline
\end{tabular}
\end{table*}

\begin{figure}[t]
    \centering
    \includegraphics[width=1.\linewidth]{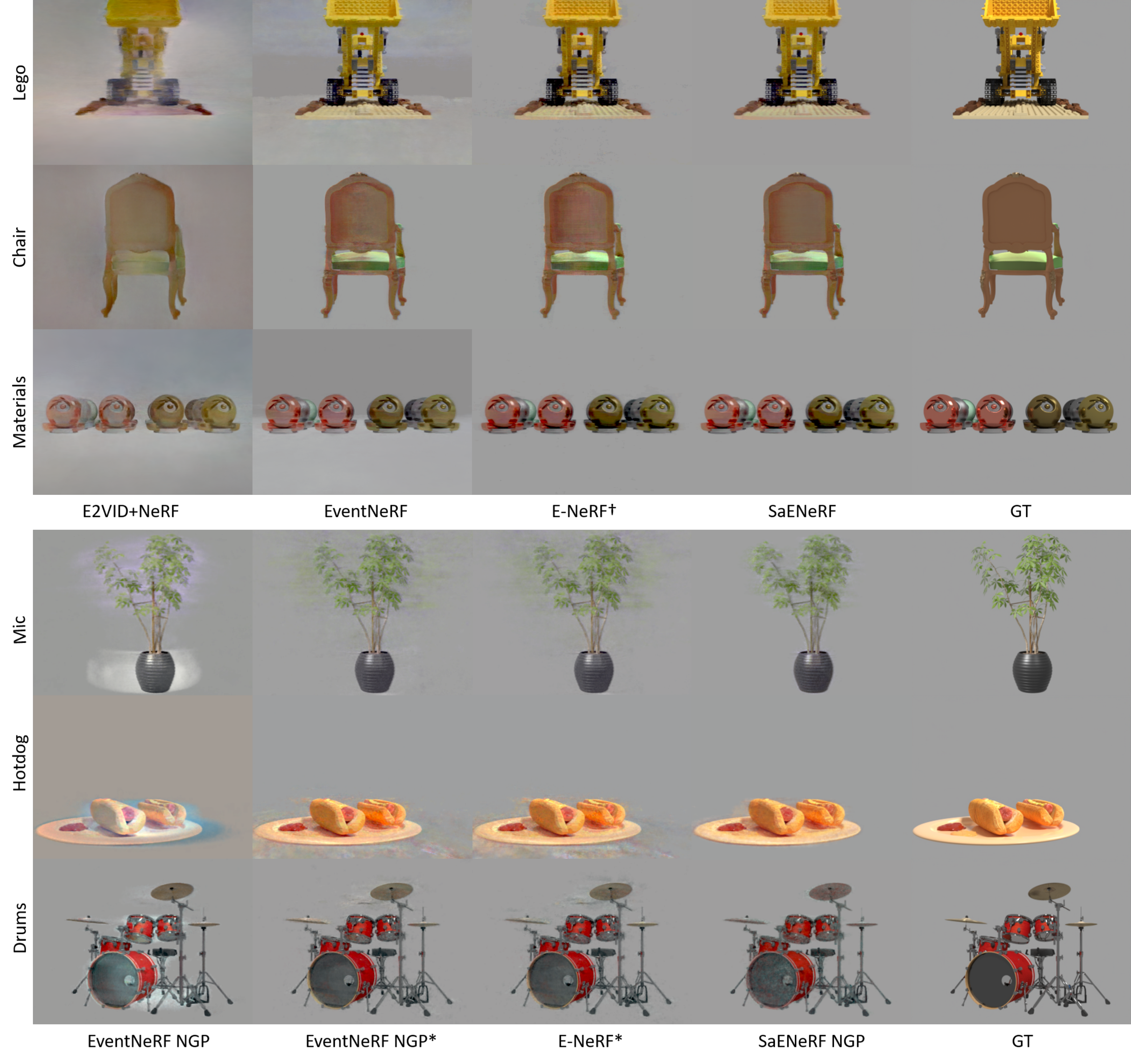}
    \caption{
        Comparison of our method with EventNeRF, E2VID+NeRF,  E-NeRF and its reimplemented real-time versions in synthetic sequences trained using only event input (no RGB).
    }
    \label{fig:syn}
\end{figure}

The quantitative results presented in Tab.\ref{tab:synthe} demonstrate that our method almost consistently outperforms both E2VID+NeRF, EventNeRF, and E-NeRF\dag{} across various scenes in terms of PSNR, SSIM, and LPIPS metrics. This indicates superior image reconstruction quality, accurate color reproduction, and precise edge sharpness, thus demonstrating the effectiveness of our approach in suppressing artifacts. 
This is particularly evident in the ``Lego" scene (Fig.\ref{fig:syn}), where other methods exhibit large or subtle areas of artifacts in the background, whereas our method achieves a clean background with sharp object edges.
Moreover, in textureless areas, such as the back of the chair in the ``Chair" scene (Fig.\ref{fig:syn}), our method reproduces colors more naturally and accurately than EventNeRF and E-NeRF\dag{} when compared to the target image.

To more fully demonstrate the excellent effect of our method in suppressing artifacts, we adopted a negative sampling ratio \cite{rudnev2023eventnerf} of 0.05 in SaENeRF NGP, E-NeRF* and EventNeRF NGP*. In contrast, EventNeRF NGP uses a negative sampling ratio of 0.1 on synthetic datasets. It's worth noting that a lower negative sampling ratio often leads to more artifacts in the background. However, even with such a low sampling ratio, our method can effectively suppress the generation of artifacts, further proving the superiority of our approach as displayed in Fig.\ref{fig:syn}. 
Specifically, EventNeRF NGP suffers from severe color artifacts, while EventNeRF NGP* and E-NeRF* show some improvement, but more complex artifacts may still appear in the background. In contrast, our method effectively suppresses the generation of these artifacts, achieving the best reconstruction results.

\begin{table*}
\caption{Quantitative comparison real-time SaENeRF NGP against EventNeRF NGP, EventNeRF NGP* and E-NeRF*.}
\label{tab:synthe_realtime}
\centering\setlength{\tabcolsep}{1.3mm}
% \begin{tabular}{c|c|c|c|c|c|c|c|c|c}
\begin{tabular}{c c c c c c c c c c c c c}
\hline
~ & \multicolumn{3}{c}{EventNeRF NGP} & \multicolumn{3}{ c}{EventNeRF NGP*} & \multicolumn{3}{ c}{E-NeRF*} & \multicolumn{3}{ c}{SaENeRF NGP} \\
% \hline
Scene & PSNR↑ & SSIM↑ & LPIPS↓ & PSNR↑ & SSIM↑ & LPIPS↓ & PSNR↑ & SSIM↑ & LPIPS↓ & PSNR↑ & SSIM↑ & LPIPS↓\\
\hline
Drums     & 26.03 & 0.91 & 0.07 & \bf28.26 & 0.91 & 0.08 & 27.99 & 0.90 & 0.10 & 28.20 & \bf0.92 & \bf0.06 \\
Lego      & 22.82 & 0.89 & 0.08 & 28.82 & \bf0.92 & 0.06 & 28.74 & 0.92 & 0.06 & \bf28.86 & \bf0.92 & \bf0.05 \\
Chair     & 27.97 & 0.94 & 0.05 & 30.48 & 0.92 & 0.10 & 30.68 & 0.93 & 0.09 & \bf31.13 & \bf0.94 & \bf0.04 \\
Ficus     & 26.77 & 0.92 & 0.12 & 30.70 & 0.91 & 0.16 & 30.44 & 0.91 & 0.15 & \bf31.61 & \bf0.94 & \bf0.05 \\
Mic       & 28.34 & 0.95 & 0.04 & 31.29 & 0.94 & 0.09 & 30.39 & 0.92 & 0.12 & \bf32.07 & \bf0.96 & \bf0.03 \\
Hotdog    & 23.99 & 0.93 & 0.10 & 30.31 & 0.92 & 0.08 & 29.80 & 0.91 & 0.08 & \bf31.90 & \bf0.95 & \bf0.04 \\
Materials & 26.05 & 0.93 & 0.07 & \bf30.36 & \bf0.95 & \bf0.05 & 30.20 & 0.94 & 0.06 & 30.20 & \bf0.95 & \bf0.05 \\
\hline
Average   & 25.99 & 0.92 & 0.07 & 30.03 & 0.92 & 0.09 & 29.75 & 0.92 & 0.09 & \bf30.57 & \bf0.94 & \bf0.05 \\
\hline
\end{tabular}
\end{table*}

Tab.\ref{tab:synthe_realtime} quantitatively demonstrates the artifact removal efficacy of our proposed method. In comparison to EventNeRF NGP* and E-NeRF*, our approach notably reduces artifact generation in the background. This is especially evident in the ``Ficus" scene, where our method achieved a significantly lower LPIPS score of 0.05 compared to the scores of 0.16 and 0.15 obtained by EventNeRF NGP* and E-NeRF* respectively. This clearly illustrates the superior performance of our method in artifact suppression.
Furthermore, when compared to EventNeRF NGP, SaENeRF NGP yields better results in terms of PSNR and SSIM, while maintaining comparable performance in LPIPS.

\begin{figure}[t] 
    \centering
    \includegraphics[width=0.87\linewidth]{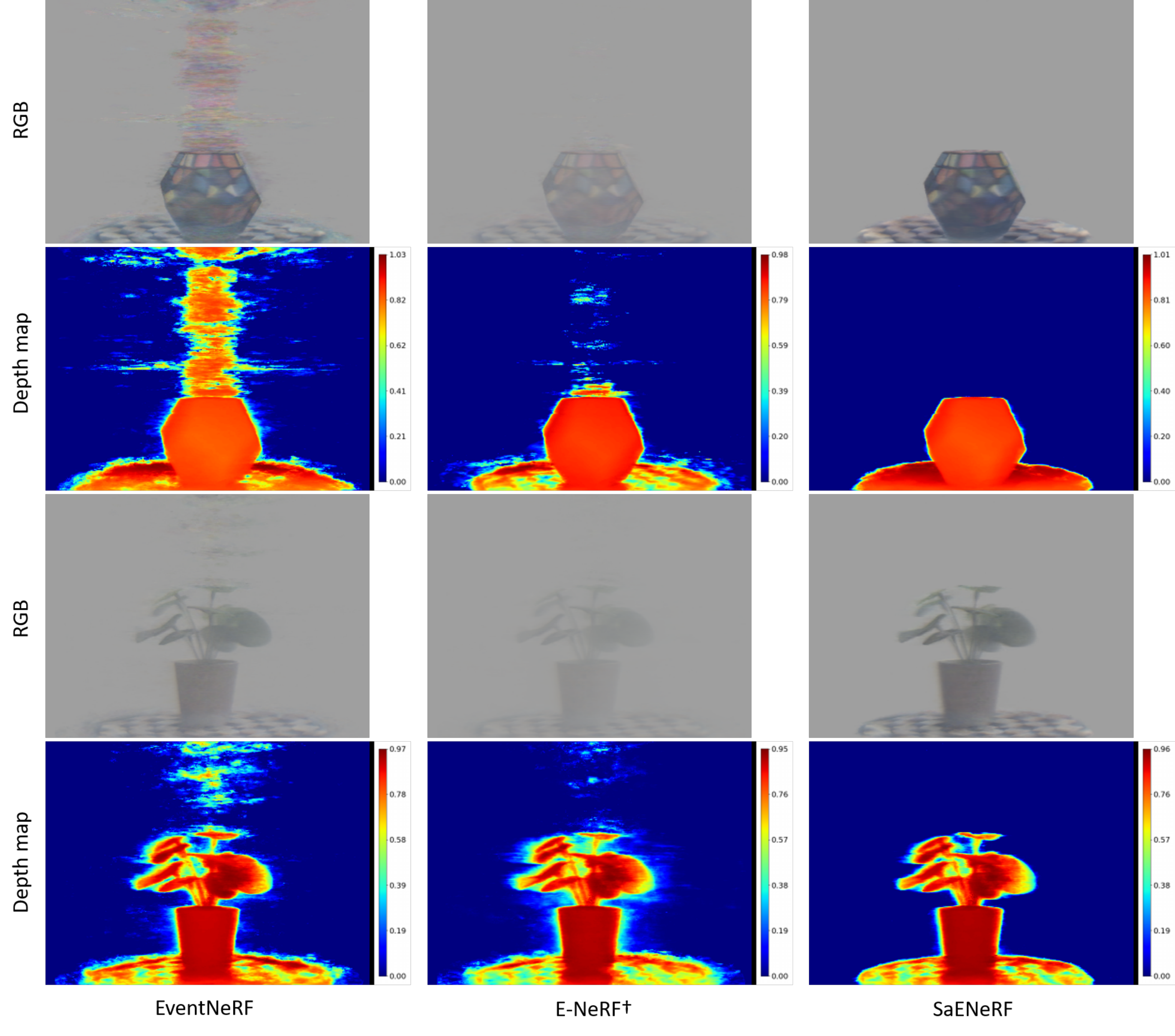}
   \caption{
        Comparison of our method with EventNeRF and E-NeRF\dag{}, presented in rendered RGB and Depth Maps in the different real sequences.
    }
    \label{fig:real}
\end{figure}

\begin{figure}[t] 
    \centering
    \includegraphics[width=1\linewidth]{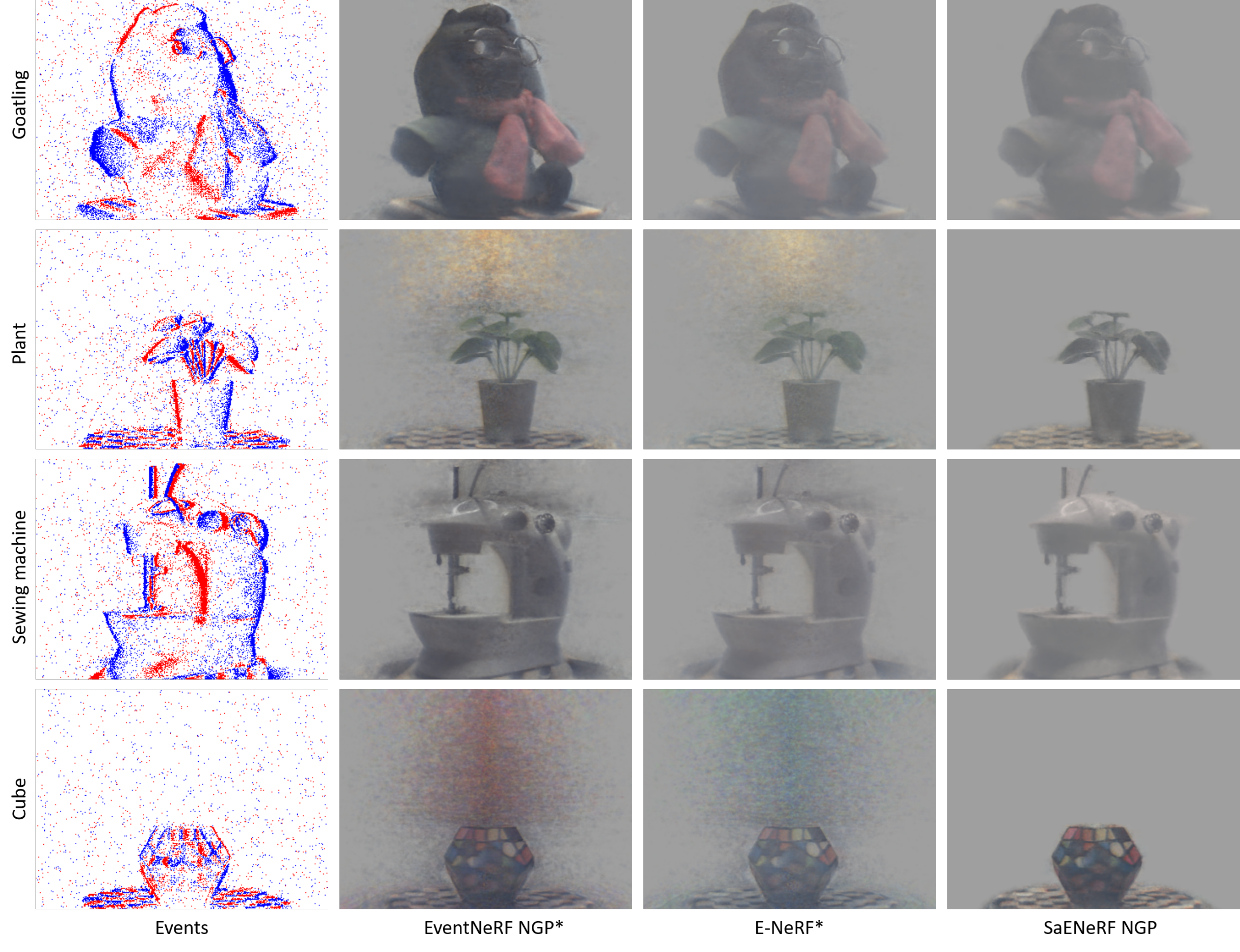}
    \caption{
        Comparison of SaENeRF NGP against EventNeRF NGP* and E-NeRF* trained for only 5000 iterations in real sequences.
    }
    \label{fig:real_real-time}
\end{figure}

\subsection{Real sequences}
\label{subsec:real}

Due to the absence of ground-truth RGB data for quantitative assessment in real sequences such as HDR or motion-blur, only qualitative evaluations are provided. To more accurately illustrate artifacts arising from the event camera's background noise in real-world environments, we avoided employing the density cropping technique \cite{rudnev2023eventnerf} previously used in the z-axis direction. Furthermore, we did not apply any manual color mapping techniques \cite{rudnev2023eventnerf} as our SaENeRF method adeptly preserves the original colors of objects while effectively suppressing artifacts.

The efficacy of artifact elimination is qualitatively evaluated, as demonstrated in Fig.\ref{fig:real} and Fig.\ref{fig:real_real-time}. 
For a more intuitive comparison, we provide rendered depth maps (see Fig.\ref{fig:real}). It is evident that the 3D reconstructions of EventNeRF and E-NeRF\dag{} exhibit numerous blurs along object edges, and EventNeRF shows annular artifacts in the center of the background. 
In contrast, our method produces reconstructions with clearer edges for the scene objects and achieves a much cleaner background, highlighting the superiority of our approach. Furthermore, the results of our real-time method in the real scenes are presented in Fig.\ref{fig:real_real-time}. It can be clearly seen that our method effectively suppresses artifacts in the background caused by noisy events while preserving fine structures such as the branches of the ``Plant" and the needle of the ``Sewing Machine".

\subsection{Ablation study}
\label{subsec:abla}

\begin{table}
    \centering
    \caption{Quantitative ablation study with different normalization and zero-events regularization components. 
    An asterisk (*) denotes the presence of 20\% event noise, while a dagger (\dag) indicates results obtained after $5\cdot10^4$ iterations.
    }
    \label{tab:abla_drums}
    \begin{tabular}{c c c c}
        \hline
         & PSNR↑ & SSIM↑ & LPIPS↓ \\
        \hline
        \text{norm+\&zero+}  & \textbf{28.02} & \textbf{0.92} & \textbf{0.07}  \\
        norm-\&zero+ & {28.00} & \textbf{0.92} & \textbf{0.07}  \\
        \text{*norm+\&zero+} & {27.88} & {\textbf{0.92}} & {\textbf{0.07}} \\
        {*norm-\&zero+} & {27.84} & {\textbf{0.92}} & {\textbf{0.07}} \\
        \text{\dag norm+\&zero+} & 26.93 & 0.90 & 0.12 \\
        {\dag norm-\&zero+} & 26.33 & 0.88 & 0.18 \\
        norm\&zero+ & 27.78 & 0.91 & 0.08  \\
        L2 norm+\&zero+ & 27.88 & 0.91 & 0.08 \\
        norm+\&zero- & 27.86 & 0.90 & 0.08 \\
        norm+\&zero & 27.64 & 0.90 & 0.08\\
        norm+ & 27.96 & 0.91 & \textbf{0.07} \\
        \hline
    \end{tabular}
\end{table}

\begin{figure}[t]
    \centering
    \includegraphics[width=.8\linewidth]{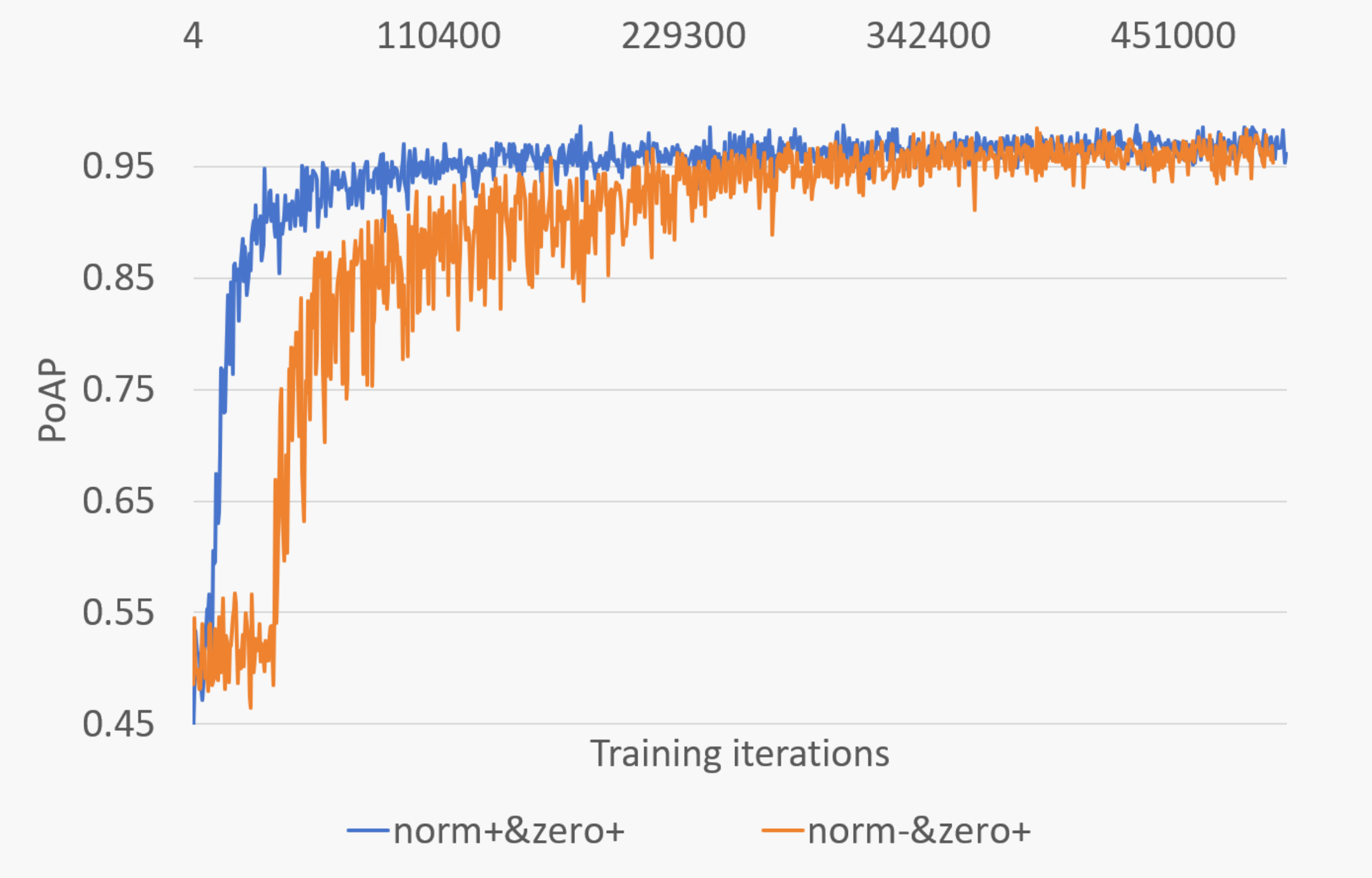}
    \caption{Ablations of different normalizations in synthetic sequence.
    }
    \label{fig:abla_drums_line}
\end{figure}

\begin{figure}[t]
    \centering
    \includegraphics[width=.8\linewidth]{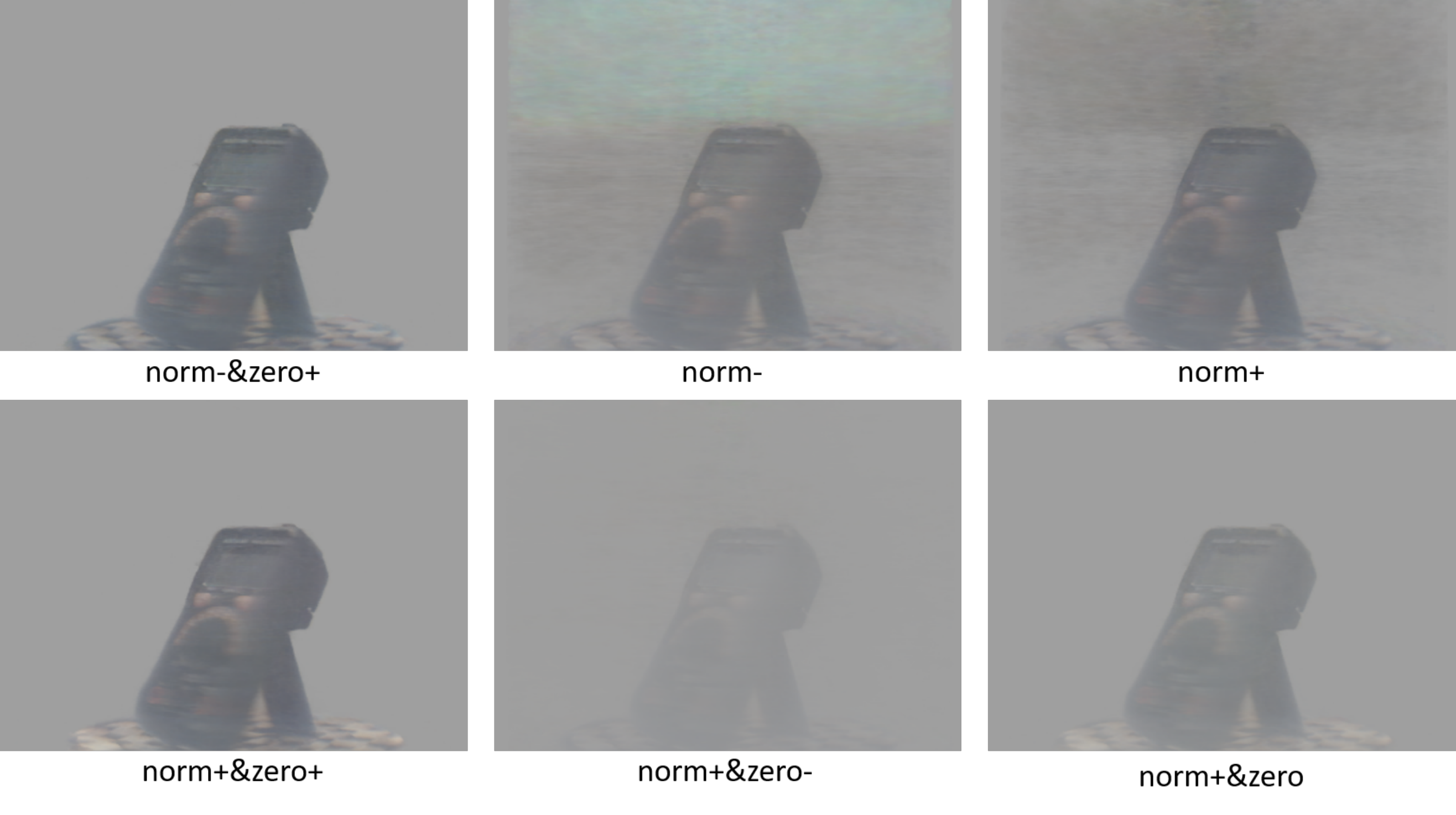}
    \caption{Ablations of different normalizations and zero-events losses in real sequence.}
    \label{fig:abla_real}
\end{figure}

We quantitatively analyzed the impact of various components in the synthetic scene of ``Drams' under different experimental conditions (see Tab.\ref{tab:abla_drums}). The results indicate that the norm+\&zero+ and norm-\&zero+ methods almost yield best reconstruction quality, both in noise-free conditions and in the presence of 20\% event noise.
However, as depicted in Fig.\ref{fig:abla_drums_line}, the norm+\&zero+ method converges significantly faster than the norm-\&zero+ method and exhibits more stable performance. Thus, the norm+\&zero+ method was chosen due to its superior results (see Tab.\ref{tab:abla_drums}).
In contrast, the norm\&zero+, norm+\&zero-, norm+\&zero and L2 norm+\&zero+ methods show slightly lower performance. The norm\&zero+ method's performance may be affected by noise introduced during normalization due to zero-events. The L2 norm+\&zero+ method's sensitivity to outliers may explain its sub-optimal results. The strict constraints on zero-events imposed by the norm+\&zero- and norm+\&zero methods result in low contrast (see Fig.\ref{fig:abla_real}).

In the real sequence, we qualitatively evaluated the effectiveness of different components in artifacts suppression. Consistent with previous findings, the reconstruction results of the norm+\&zero+ and norm-\&zero+ methods are similar, with norm+\&zero and norm+\&zero- performing slightly worse.
Significant degradation is mainly observed in the absence of zero-events regularization. Interestingly, the norm+ method produces fewer artifacts than the norm- method, suggesting that norm+ also has artifacts suppression capabilities.

\section{Conclusion}
\label{conclusion}

We present SaENeRF, which, for static scenes captured using moving event cameras, can directly perform 3D-consistent, dense, and photorealistic NeRF reconstruction from asynchronous event streams.
By utilizing novel normalization technique and zero-event regularization losses, we have effectively reduced artifacts, resulting in high-quality 3D reconstructions. 
This is confirmed by both quantitative and qualitative experiments on synthetic and real event sequences. 
However, although our method has the capability to effectively suppress artifacts, it is unable to achieve their complete elimination. In addition, its reliance on known poses could limits its practical applications. 
Therefore, exploring more effective artifact elimination methods based on events and investigating event-based implicit neural SLAM in pose-unknown scenarios with normalized reconstruction losses are valuable directions.

\section*{Acknowledgements}
This study was supported by the Natural Science Foundation of Sichuan (Grant No.2024NSFSC1470 and 24NSFSC3404), National Natural ScienceFoundation of China (Grant No.62206188), National Major Scientific In-struments and Equipments Development Project of National Natural ScienceFoundation of China (Grant No. 62427820) and Fundamental Research Fundsfor the Central Universities (Grant No. 1082204112364).

\bibliographystyle{IEEEtran}
\bibliography{main}

\end{document}